\documentclass[conference,compsoc]{IEEEtran}
\usepackage{booktabs}
\usepackage{multirow}
\newcommand{\squeeze}{\vspace{-2.5mm}} 
\usepackage[centerlast,small,sc]{caption}
\usepackage[usenames, dvipsnames]{color} 
\usepackage{amsmath,amssymb}
\definecolor{purple}{rgb}{0.5, 0.0, 0.5}
\definecolor{orange}{rgb}{1, 0.65, 0}
\definecolor{lightgreen}{rgb}{0.68, 1, 0.18}
\definecolor{darkgreen}{rgb}{0.09, 0.32, 0.24}
\definecolor{darkred}{rgb}{0.6, 0, 0}
\definecolor{brown}{rgb}{0.64, 0.16, 0.16}
\iffalse 
  \newcommand{\holger}[1]{\noindent}
  \newcommand{\oscar}[1]{\noindent}
  \newcommand{\yiluan}[1]{\noindent}
  \newcommand{\done}[1]{\noindent}
  \newcommand{\todo}[1]{\noindent}
\else
  \newcommand{\holger}[1]{\textcolor{blue}{\bf [HC: #1]}}
  
  \newcommand{\oscar}[1]{\textcolor{orange}{\bf [OB: #1]}}
  \newcommand{\yiluan}[1]{\textcolor{purple}{\bf [YG: #1]}}
  \newcommand{\done}[1]{\textcolor{darkgreen}{\bf [Done: #1]}}
  \newcommand{\todo}[1]{\textcolor{red}{\bf [Todo: #1]}}
\fi

\ifCLASSOPTIONcompsoc
  \usepackage[nocompress]{cite}
\else
  \usepackage{cite}
\fi

\ifCLASSINFOpdf
  \usepackage[pdftex]{graphicx}
  \graphicspath{{figures/}}
  \DeclareGraphicsExtensions{.pdf,.jpeg,.png}
\else
\fi

\begin{document}
%
\title{The efficacy of Neural Planning Metrics: A meta-analysis of PKL on nuScenes \vspace{-4mm}}

\author{\IEEEauthorblockN{Yiluan Guo *}
\and
\IEEEauthorblockN{Holger Caesar*}
\IEEEauthorblockA{\\Motional*}
\and
\IEEEauthorblockN{Oscar Beijbom*}
\and
\IEEEauthorblockN{Jonah Philion$^\dagger$}
\IEEEauthorblockA{\\NVIDIA$^\dagger$}
\and
\IEEEauthorblockN{Sanja Fidler$^\dagger$}
}

\maketitle

\begin{abstract}
A high-performing object detection system plays a crucial role in autonomous driving (AD). The performance, typically evaluated in terms of mean Average Precision, does not take into account orientation and distance of the actors in the scene, which are important for the safe AD. It also ignores environmental context.
Recently, \cite{philion2020learning} proposed a neural planning metric (PKL), based on the KL divergence of a planner's trajectory and the groundtruth route, to accommodate these requirements. In this paper, we use this neural planning metric to score all submissions of the nuScenes detection challenge and analyze the results. We find that while somewhat correlated with mAP, the PKL metric shows different behavior to increased traffic density, ego velocity, road curvature and intersections. Finally, we propose ideas to extend the neural planning metric.
\end{abstract}

\squeeze
\section{Introduction}
\squeeze

In recent years we have witnessed rapid progress in the field of AD. An essential task in AD perception is object detection. Object detection is typically evaluated by mean Average Precision~(mAP), which takes into account false positives and false negative detections at different classification thresholds. However, this metric may be too simplistic for the complexity of the AD task, which requires estimate object orientations and velocities and where importance of an error is determined by the context. E.g. benchmarks~\cite{kitti,nuscenes,waymo_open_dataset} give the same importance to two cars regardless of their ego distance.

Several metrics have been designed to be more relevant to the AD task. For instance, the nuScenes detection score (NDS)~\cite{nuscenes} is defined as the weighted sum of mAP and several true positive metrics such as translation, orientation, rotation, attribute and velocity errors.
\cite{waymo_open_dataset} proposed a novel metric that weights the true positives in mAP by their heading accuracy.
These metrics evaluate the detection system more comprehensively, but they do not consider the interactions between objects.

Recently a new metric based on a pretrained neural planner~(PKL) was proposed in \cite{philion2020learning}. The planner is conditioned on the detected bounding boxes and the local semantic map. The planner learns from a large amount of human driving data how navigate the scenes.
The authors of~\cite{philion2020learning} have conducted an insightful analysis of a selected object detection method~\cite{megvii}.
In this work, we perform a meta analysis on the variation of the PKL metric based on traffic density, ego velocity, road curvature and intersections, using the nuScenes dataset.

\squeeze
\section{Methodology}
\label{sec:methodology}
\squeeze
We re-evaluate the test set submissions of the nuScenes~\cite{nuscenes} detection challenge.
All results are aggregated on a scene-level or beyond to preserve the integrity of the test set annotations.
The nuScenes submissions contain the maximum a posteriori class name and score for each predicted bounding box.
PKL does not take into account the detection score, but rather operates on the set of positives after score thresholding.
To perform a fair comparison between different submissions, we compute the ``optimal confidence threshold'' of a detector for each class.
The optimal confidence threshold among all box scores is the threshold that achieves the highest F1 score. 
We show in Section~\ref{sec:confidence_treshold} that this choice is indeed close to the threshold that achieves the top performing PKL value.
We then remove the predicted boxes for the specified class that are below the optimal confidence threshold.
Finally, we compute PKL and other metrics for each sample (keyframe) in the nuScenes dataset and perform various analyses on the data.

We focus our analysis on three  methods: A vision-only method: MonoDIS~\cite{monodis} and two lidar-only methods MEGVII~\cite{megvii} and Noah.

\squeeze
\section{Experiments}
\squeeze
In this section we present results and analysis on the nuScenes detection challenge.

\squeeze
\subsection{Metric comparison}
\begin{figure}
 \centering 
 \vspace{-7mm}
 \includegraphics[width=0.8\linewidth]{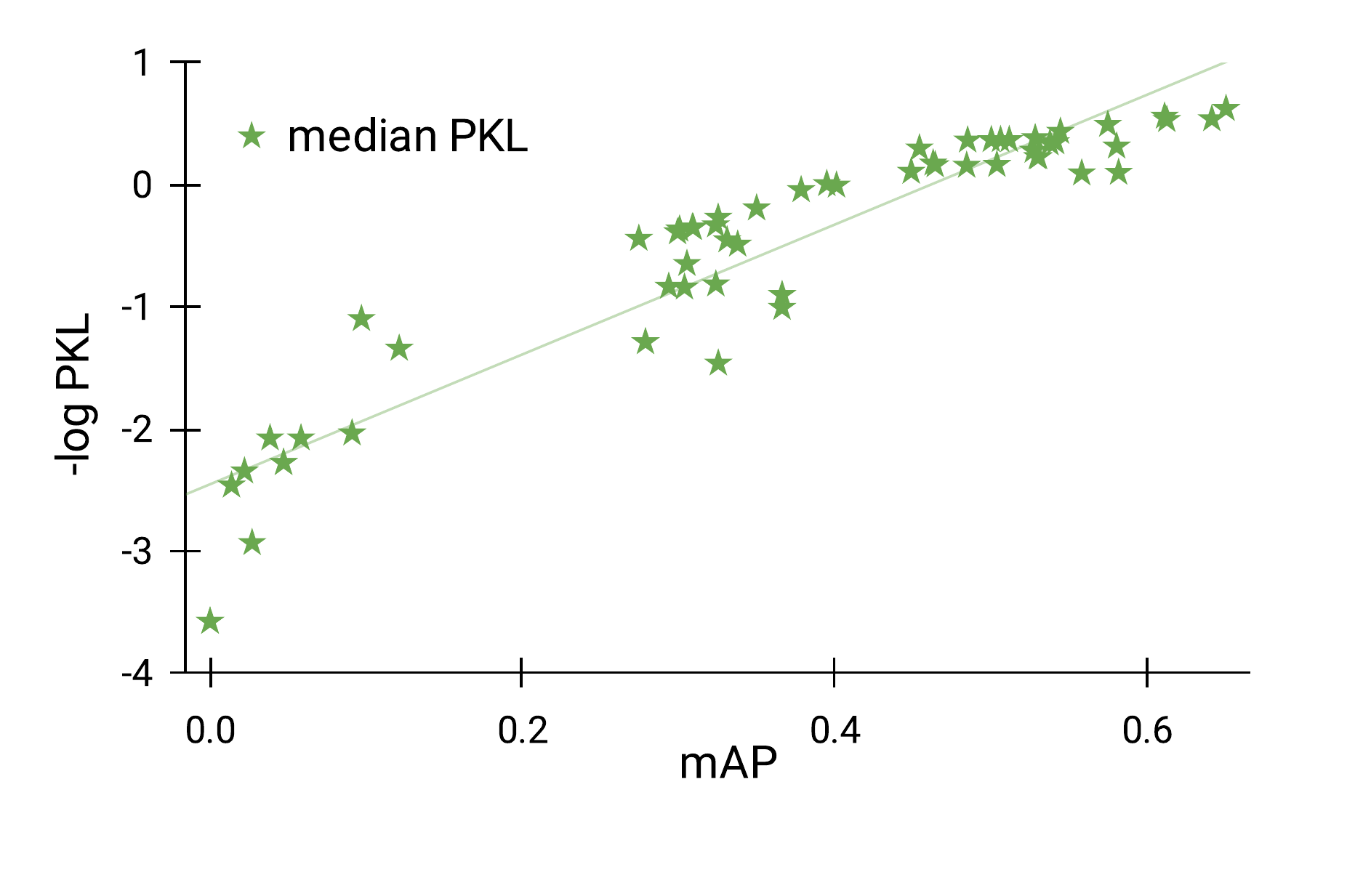}
 \vspace{-4mm}
 \caption{mAP vs. median PKL. Note that we plot the negative log of PKL.}
 \vspace{-1mm}
 \label{metric_comparsion}
\end{figure}

\squeeze
We plot mAP against median PKL scores for all submitted methods in the nuScenes detection challenge, as shown in Figure \ref{metric_comparsion}. Since NDS and mAP are strongly correlated, we ignore NDS for simplicity. 
Unless specified otherwise, we use the median of per sample PKL values (``median PKL'') rather than the mean, to reduce the impact of outliers.
The inverted log scale is applied to magnify the variation of PKL.

As Philion \textit{et al.}~\cite{philion2020learning} have shown on the val set, dropping the detections leads to a mean PKL of 54.63. Using the same pretrained planner, we observe that two submissions with an mAP of 0 have a mean PKL of 44.75 on the test set.

As expected, PKL scores are consistent with mAP in general. 
We observe a strong Spearman rank correlation of 0.93. For the competitive methods (mAP $>$ 0.45) rank correlation decreases to only 0.47. This means that when using PKL as a benchmark for AD, the ranking of the top methods would change significantly compared to mAP.

\squeeze
\subsection{Traffic density}
\label{sec:traffic_density}
\begin{figure}%
    \centering
    \includegraphics[width=0.8\linewidth]{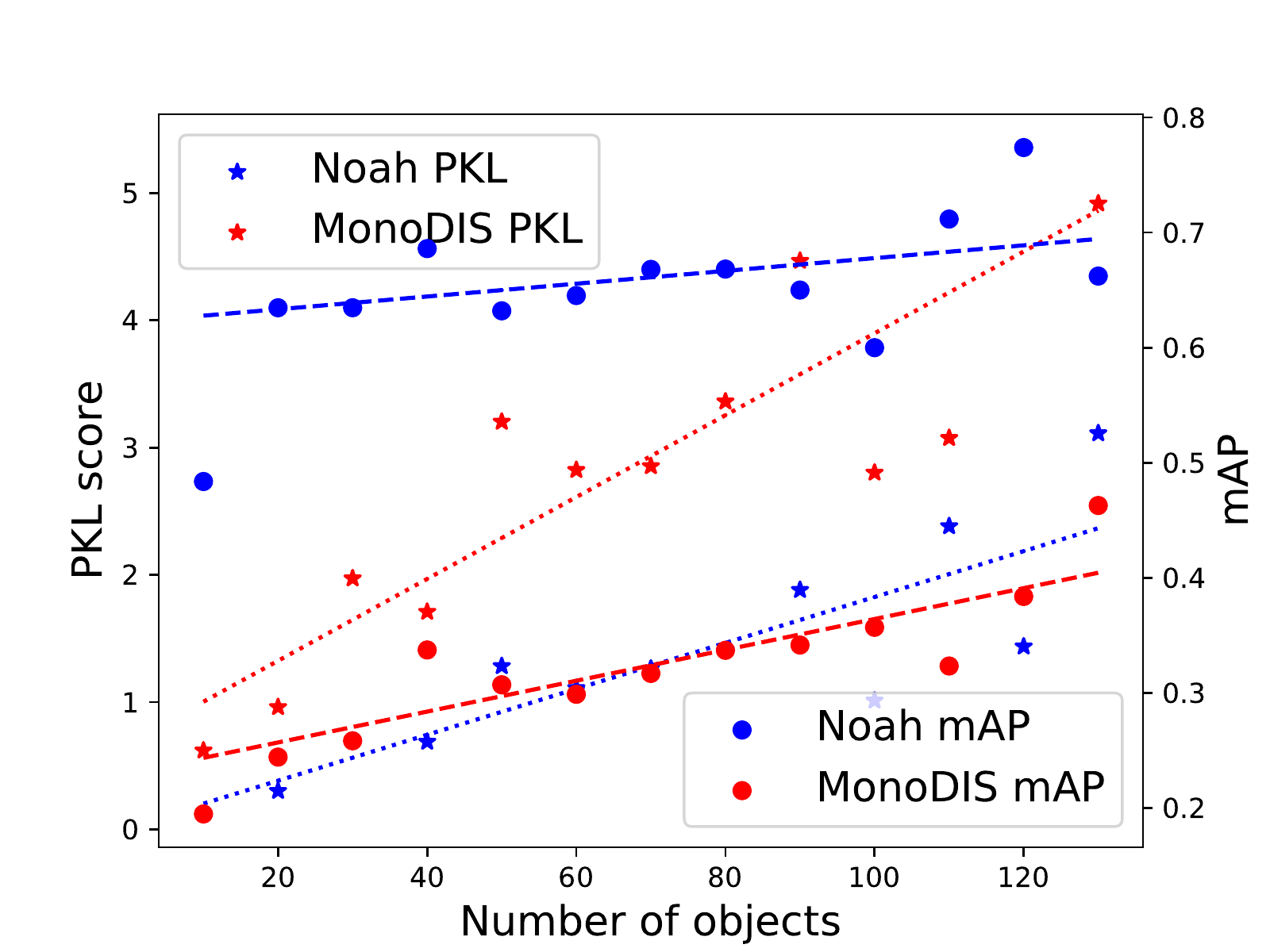}
    \vspace{+2mm}
    \caption{Number of objects vs. PKL$\downarrow$ and mAP$\uparrow$ metrics for the Noah and MonoDIS submissions. The number of objects is computed per sample and averaged over bins of uniform width.}
    \vspace{-3mm}
    \label{density}
\end{figure}

\squeeze
 We analyze the impact of traffic density, approximated by the number of objects per sample, on the detection performance in terms of PKL and mAP (Figure~\ref{density}).
Noah and MEGVII are shown similar performance here; we do not show the latter for clarity.
When more objects are in the scene, PKL deteriorates, while mAP improves.
For MonoDIS the deterioration in PKL is much more pronounced than for Noah, possibly due to monocular methods producing noisier predictions along the depth dimension, causing the planner to adjust its trajectory.
Numeric results are provided in Table~\ref{num_objects}.

\squeeze
\subsection{Intersections}
\label{sec:intersections}
\squeeze
In this section we investigate the hypothesis that the planning task is particularly difficult around intersections where small errors in perception are more likely have a large affect on downstream planning. We therefore expect PKL to penalize detection mistakes more strongly near intersections. To evaluate this intuition quantitatively, we query for each sample in nuScenes whether its ego pose falls on a ``road segment'' with \emph{is\_intersection} set to True. In total this is true for 30\% of the samples. 
We compare the performance on intersection samples with the performance on all test set samples.
We observe that the median PKL for intersections deteriorates by +0.06 (0.71) for ``Noah'', by +0.11 (0.88) for ``MEGVII''~\cite{megvii} and by +0.18 (1.97) for ``MonoDIS''~\cite{monodis}.
The mAP of these three methods, as expected, does not change significantly, since the perception task is not harder at intersections.

\squeeze
\subsection{Ego vehicle velocity}
\squeeze
In \cite{philion2020learning} the authors demonstrate that PKL deteriorates for predicted objects with a higher speed.
In Table~\ref{speed} we analyze the PKL on individual samples with different \emph{ego vehicle} speeds.
The samples are binned by their speed and we accumulate the PKL and mAP scores within each bin.
We observe that PKL varies significantly with speed, whereas mAP varies less.
In particular, PKL achieves its best performance for stationary ego vehicles, since the planning task often becomes trivial in stationary scenes.
For mAP the trend is less clear.
mAP depends on the detector being evaluated and is likely an artifact of the data distribution at different speeds, such as a freeway versus a parking lot.

\squeeze
\subsection{Rain and night}
\squeeze
We further analyze the PKL scores for two challenging types of scenes: rain and night (Table \ref{night_rain}).
For mAP we see that the performance deteriorates on both rain and night.
The decrease in performance is more pronounced for night scenes, in particular for the monocular MonoDIS method.
Surprisingly for PKL we see improvements for night data and deterioration for rain data.
Possible explanations may be the different data and object distribution at night. Whereas busy rush hour scenes may pose challenges for the planner, at night there are fewer objects (in particular pedestrians), which may simplify the planning task.

\begin{table}[]
\begin{tabular}{cllllll}
\hline
\multirow{2}{*}{Scenes} & \multicolumn{2}{c}{MEGVII} & \multicolumn{2}{c}{MonoDIS} & \multicolumn{2}{c}{Noah} \\ \cline{2-7} 
 & PKL$\downarrow$ & mAP$\uparrow$ & PKL$\downarrow$ & mAP$\uparrow$ & PKL$\downarrow$ & mAP$\uparrow$ \\ \hline
All scenes & 0.77 & \textbf{0.53} & 1.79 & \textbf{0.30} & 0.65 & \textbf{0.65} \\
Rain & 1.12 & 0.42 & 2.52 & 0.28 & 0.98 & 0.56\\
Night & \textbf{0.30} & 0.37 & \textbf{1.22} & 0.08 & \textbf{0.25} & 0.39 \\ \hline
\end{tabular}
\caption{PKL and mAP for rain and night scenes. Best performances are highlighted. Increases and decreases versus the baseline are shown as up and down arrows. Note that an increase in PKL means a deterioration in performance and vice versa.}
\vspace{-4mm}
\label{night_rain}
\end{table}

\squeeze
\section{Conclusion}
\squeeze
In this work, we conducted a meta analysis on a recently proposed neural planning metric PKL. PKL is consistent with mAP on a macro level. However the behavior for PKL and mAP differs for selected criteria such as increased traffic density, intersections, high ego vehicle velocity, as well as rain and night data. 
To improve the PKL metric we propose the following extensions:
(1) Include past sensor data and object trajectories. This data may be beneficial to estimate the future trajectory of an object, which would in turn improve the planner and the metric. 
(2) Include high-level planning goals to reduce ambiguity and improve the metric, such as the navigation baseline routes in nuScenes. Without these, the planner may not know whether to turn left or right at an intersection.


\clearpage
\bibliographystyle{IEEEtran}
\bibliography{reference.bib}

\appendix

\begin{table}[]
\centering
\begin{tabular}{@{}cccccccc@{}}
\toprule
\multirow{2}{*}{\begin{tabular}[c]{@{}c@{}}Speed \\ range\end{tabular}} & \multirow{2}{*}{\begin{tabular}[c]{@{}c@{}}Number of \\ samples\end{tabular}} & \multicolumn{2}{c}{MEGVII}    & \multicolumn{2}{c}{MonoDIS}   & \multicolumn{2}{c}{Noah}      \\ \cmidrule(l){3-8} 
                                                                        &                                                                               & PKL$\downarrow$           & mAP$\uparrow$           & PKL$\downarrow$           & mAP$\uparrow$           & PKL$\downarrow$           & mAP$\uparrow$           \\ \midrule
All scenes                                                              & 6008                                                                          & 0.77          & 0.53          & 1.79          & 0.30          & 0.65          & 0.65          \\
0                                                                       & 654                                                                           & \textbf{0.13} & 0.49          & \textbf{0.53} & \textbf{0.37} & \textbf{0.14} & 0.62          \\
(0, 2{]}                                                                & 374                                                                           & 1.35          & 0.51          & 3.38          & 0.36          & 1.15          & 0.65          \\
(2, 4{]}                                                                & 740                                                                           & 1.24          & 0.49          & 3.58          & 0.31          & 1.18          & 0.63          \\
(4, 6{]}                                                                & 1340                                                                          & 1.04          & 0.52          & 2.42          & 0.28          & 0.84          & 0.64          \\
(6, 8{]}                                                                & 1403                                                                          & 1.03          & 0.53          & 1.97          & 0.30          & 0.83          & 0.65          \\
(8, 10{]}                                                               & 1234                                                                          & 0.44          & 0.57          & 1.08          & 0.31          & 0.37          & \textbf{0.70} \\
(10, 12{]}                                                              & 263                                                                           & 0.37          & \textbf{0.58} & 0.76          & 0.30          & 0.26          & \textbf{0.70} \\ \bottomrule
\end{tabular}
\caption{PKL and mAP for different ego speed values. Best performances are highlighted.
}
\label{speed}
\end{table}

\subsection{Confidence threshold}
\label{sec:confidence_treshold}
In Section~\ref{sec:methodology} we described how during evaluation we threshold the predicted boxes to create a set of positives and negatives.
This confidence threshold is determined by the highest F1 score.
The choice is independent of PKL itself and hence we investigate if it is suitable for the PKL metric.
We uniformly sample thresholds between 0.0 and 0.9 for the ``car" class, while keeping them constant for other classes.

From Table \ref{confidence_threshold}, we can see the PKL performances improve when the threshold increases from 0 to an optimum, due to the removal of false positives. 
As the threshold increases further, false negatives dominate and PKL deteriorates.
The best performance is very close to the one with highest F1 score, which suggests that this choice is indeed optimal for minimizing PKL.

\begin{table}[]
\centering
\begin{tabular}{@{}cccc@{}}
\toprule
\begin{tabular}[c]{@{}c@{}}Confidence threshold \\ for car\end{tabular} & MEGVII & MonoDIS & Noah \\ \midrule
\begin{tabular}[c]{@{}c@{}}Threshold from \\ highest F1\end{tabular} & 0.77 & 1.79 & 0.65 \\ \midrule
0 & 1.35 & 4.23 & 2.02 \\
0.1 & 1.35 & 2.30 & 2.02 \\
0.2 & 0.85 & 1.79 & 0.93 \\
0.3 & \textbf{0.77} & \textbf{1.78} & 0.71 \\
0.4 & 0.80 & 1.87 & 0.66 \\
0.5 & 0.90 & 2.05 & \textbf{0.64} \\
0.6 & 1.02 & 2.57 & 0.67 \\
0.7 & 1.23 & 3.94 & 0.82 \\
0.8 & 1.68 & 5.45 & 1.61 \\
0.9 & 2.61 & 5.73 & 3.60 \\ \bottomrule
\end{tabular}
\caption{PKL metrics with different confidence threshold values for ``car" class. Best performances are highlighted.}
\label{confidence_threshold}
\end{table}

\begin{table}[]
\begin{tabular}{@{}cccccccc@{}}
\toprule
\multirow{2}{*}{\begin{tabular}[c]{@{}c@{}}Number of \\ objects\end{tabular}} & \multirow{2}{*}{\begin{tabular}[c]{@{}c@{}}Number of \\ samples\end{tabular}} & \multicolumn{2}{c}{MEGVII} & \multicolumn{2}{c}{MonoDIS} & \multicolumn{2}{c}{Noah} \\ \cmidrule(l){3-8} 
 &  & PKL$\downarrow$ & mAP$\uparrow$ & PKL$\downarrow$ & mAP$\uparrow$ & PKL$\downarrow$ & mAP$\uparrow$ \\ \midrule
All scenes & 6008 & \multicolumn{1}{l}{0.77} & \multicolumn{1}{l}{0.53} & \multicolumn{1}{l}{1.79} & \multicolumn{1}{l}{0.30} & \multicolumn{1}{l}{0.65} & \multicolumn{1}{l}{0.65} \\
(0, 10{]} & 865 & \textbf{0.20} & 0.38 & \textbf{0.62} & 0.19 & \textbf{0.12} & 0.48 \\
(10, 20{]} & 1147 & 0.38 & 0.51 & 0.96 & 0.24 & 0.30 & 0.63 \\
(20, 30{]} & 1001 & 0.83 & 0.51 & 1.97 & 0.26 & 0.69 & 0.63 \\
(30, 40{]} & 972 & 0.78 & 0.56 & 1.71 & 0.34 & 0.69 & 0.69 \\
(40, 50{]} & 698 & 1.45 & 0.49 & 3.20 & 0.31 & 1.28 & 0.63 \\
(50, 60{]} & 506 & 1.37 & 0.51 & 2.82 & 0.30 & 1.11 & 0.65 \\
(60, 70{]} & 318 & 1.53 & 0.54 & 2.85 & 0.32 & 1.26 & 0.67 \\
(70, 80{]} & 200 & 1.37 & 0.55 & 3.36 & 0.34 & 1.41 & 0.67 \\
(80, 90{]} & 110 & 1.82 & 0.52 & 4.46 & 0.34 & 1.88 & 0.65 \\
(90, 100{]} & 83 & 1.98 & 0.50 & 2.80 & 0.36 & 1.01 & 0.60 \\
(100, 110{]} & 42 & 1.98 & \textbf{0.58} & 3.07 & 0.32 & 2.38 & 0.71 \\
(110, 120{]} & 38 & 3.81 & \textbf{0.58} & 5.36 & 0.38 & 1.44 & \textbf{0.77} \\
(120, 130{]} & 24 & 6.19 & 0.54 & 4.92 & \textbf{0.46} & 3.11 & 0.66 \\ \bottomrule
\end{tabular}
\caption{PKL and mAP for different number of objects in the test samples. Best performances are highlighted.}
\label{num_objects}
\end{table}

\subsection{Simulated congestion - An upper bound}
\label{sec:simulated_congestion}

In Section~\ref{sec:traffic_density} we observed that PKL deteriorates for scenes with a larger traffic density.
Here we analyze how PKL is affected by a large number of hallucinated objects (false positives) at different distances to the ego vehicle.
We simulate a set of car detections that are located uniformly on a circle centered at the ego vehicle.
The results are shown in Table \ref{congestion}.
As expected the PKL is very high for congestion close to the ego vehicle, as the planner struggles to plan the trajectory.
These numbers form a possible upper bound for the PKL metric on the nuScenes dataset.
Surprising is that PKL does not always increase monotonously for more car false positives.
It is possible that at modest congestion the planner tries to find a way through the traffic, whereas for extreme congestion it does not plan to move forward.

\begin{table}[]
\centering
\begin{tabular}{@{}ccccc@{}}
\toprule
\multirow{2}{*}{Number of Cars} & \multicolumn{4}{c}{Distance to ego vehicle (m)} \\
 & 5 & 10 & 15 & 20 \\ \midrule
5 & 199.0 & 59.2 & 24.9 & 15.9 \\
10 & \textbf{216.7} & 161.6 & 45.3 & 22.2 \\
20 & 190.6 & \textbf{196.3} & 71.5 & 39.2 \\
30 & 179.1 & 195.4 & \textbf{124.5} & \textbf{48.3} \\ \bottomrule
\end{tabular}
\caption{PKL metrics for a circle of cars around the ego vehicle. Highest PKL scores are highlighted, indicating worst performances.}
\label{congestion}
\end{table}

\subsection{Road curvature}
In Section~\ref{sec:intersections} we showed that PKL deteriorates at intersections. 
Here we investigate how road curvature affects PKL.
We compute the curvatures of all samples based on the ego pose at each timestamp in the scene.
The ego poses are smoothed with a Gaussian filter to reduce the impact of localization errors.
We drop samples with curvatures larger than 0.1 as outliers. 
Finally, bin the samples by their curvature we compute mAP and PKL for all samples in a bin.
The results are shown in Table~\ref{curvature}.
For samples with very low curvature the PKL is usually higher.
Beyond that there seems to be no clear tendency between curvature and PKL or mAP.
Note that there are few samples with a high curvature. This makes the mAP computation particularly noisy in case of missing classes.

\begin{table}[]
\centering
\begin{tabular}{@{}ccllllll@{}}
\toprule
\multirow{2}{*}{\begin{tabular}[c]{@{}c@{}}Curvature \\ range\end{tabular}} & \multirow{2}{*}{\begin{tabular}[c]{@{}c@{}}Number of \\ samples\end{tabular}} & \multicolumn{2}{c}{MEGVII}                        & \multicolumn{2}{c}{MonoDIS}                       & \multicolumn{2}{c}{Noah}                          \\ \cmidrule(l){3-8} 
                                                                            &                                                                               & \multicolumn{1}{c}{PKL$\downarrow$} & \multicolumn{1}{c}{mAP$\uparrow$} & \multicolumn{1}{c}{PKL$\downarrow$} & \multicolumn{1}{c}{mAP$\uparrow$} & \multicolumn{1}{c}{PKL$\downarrow$} & \multicolumn{1}{c}{mAP$\uparrow$} \\ \midrule
All scenes                                                                  & 6008                                                                          & 0.77                    & 0.53                    & 1.79                    & 0.30                    & 0.65                    & 0.65                    \\
0                                                                           & 1779                                                                          & 0.99                    & 0.52                    & 2.20                    & 0.28                    & 0.87                    & 0.63                    \\
(0, 0.02{]}                                                                 & 2774                                                                          & 0.85                    & \textbf{0.55}           & 1.89                    & 0.30                    & 0.68                    & 0.67                    \\
(0.02, 0.04{]}                                                              & 274                                                                           & \textbf{0.76}           & 0.51                    & 2.43                    & 0.29                    & \textbf{0.62}           & 0.64                    \\
(0.04, 0.06{]}                                                              & 127                                                                           & 0.85                    & 0.48                    & 2.28                    & 0.31                    & 0.82                    & 0.64                    \\
(0.06, 0.08{]}                                                              & 97                                                                            & 0.77                    & 0.52                    & 2.62                    & 0.35                    & 0.74                    & \textbf{0.69}           \\
(0.08, 0.1{]}                                                               & 40                                                                            & 0.97                    & 0.47                    & \textbf{1.72}           & \textbf{0.36}           & 0.63                    & 0.64                    \\ \bottomrule
\end{tabular}
\caption{PKL and mAP for different road curvatures. Best performances are highlighted.}
\label{curvature}
\end{table}


\end{document}